\pdfoutput=1
\documentclass[11pt]{article}

\usepackage[final]{acl}

\usepackage{times}
\usepackage{latexsym}
\usepackage[T1]{fontenc}
\usepackage[utf8]{inputenc}
\usepackage{microtype}
\usepackage{inconsolata}
\usepackage{graphicx}
\usepackage{booktabs}
\usepackage{amsmath}
\usepackage{xcolor}
\usepackage{url}
\usepackage{tikz}
\usetikzlibrary{positioning,arrows.meta,fit}

\definecolor{seqlight}{HTML}{CDE2FB}
\definecolor{seqdark}{HTML}{184F95}
\definecolor{accentaqua}{HTML}{1BAF7A}
\definecolor{inkgray}{HTML}{52514E}

\graphicspath{{figures/}}
\title{Prioritize, Don't Decide: \\
What We Learned Auditing a Deployed Human-in-the-Loop CV Sorter}

\author{
  \textbf{Hadi Mohammadi}\textsuperscript{1,2},
  \textbf{Jeroen Sparla}\textsuperscript{1},
  \textbf{Alexandros Paliouras}\textsuperscript{2},
  \textbf{Georg Krempl}\textsuperscript{3},
  \textbf{Robert A. Bagheri}\textsuperscript{2}
\\
  \textsuperscript{1}AcademicTransfer, Utrecht, The Netherlands
\\
  \textsuperscript{2}Department of Methodology, Statistics \& Data Science,
  Utrecht University, The Netherlands
\\
  \textsuperscript{3}Department of Information and Computing Sciences,
  Utrecht University, The Netherlands
\\
  \texttt{mohammadi@academictransfer.nl}
}

\begin{document}
\maketitle

\begin{abstract}
A single academic vacancy can draw hundreds of applications, and the first
screening round consumes most of a committee's time. We describe a CV Priority
Sorter deployed at a national platform for academic vacancies and integrated
with its vacancy database. It proposes a reading order for a pile of CVs and
leaves every decision to the committee, pairing a frozen multilingual
transformer encoder with a per-session linear model that retrains after each
label and is discarded when the session ends. It works: on a pool of 300
synthetic CVs, a committee reading in the tool's order reaches 95\% of the
relevant candidates after half the pile, where an arbitrary order needs
almost all of it. This paper is about what we found when we audited that
claim, rather than about the system itself. Three findings changed the
product. The pool turned out to be sortable by document length alone, which
makes precision-style metrics on it uninformative; regenerating it under a
shared length budget removes that shortcut and leaves the reading-depth
result standing. The four-level agreement
score such systems report, ours included, is a binary probability cut at fixed
thresholds, so it follows the classifier's confidence rather than its ranking
and falls as labels accumulate. And uncertainty sampling, the obvious way to
implement a ``suggest the next CV'' button, leaves model quality intact but
raises the committee's total reading load above the no-label baseline, because
the CVs it asks about are the ones that would otherwise be skipped. A fourth
result shows where to look next: a hosted LLM embedding model lifts zero-label
ordinal discrimination from 0.66 to 0.80 under length control, then stops
mattering once the session model has two dozen labels; a local cross-encoder
matches that cold-start gain without sending data out. We report the fixes each
finding forced.
\end{abstract}

\section{Introduction}

\begin{figure}[!t]
\centering
\includegraphics[width=\columnwidth]{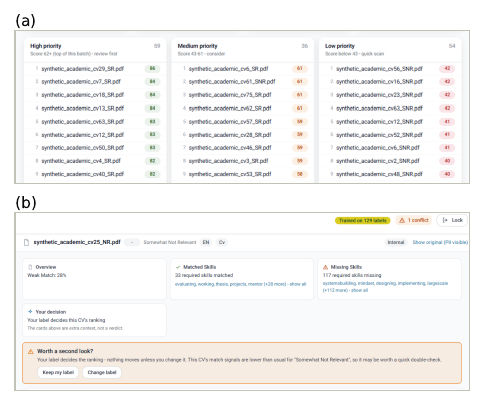}
\caption{The tool as the committee sees it. (a) The pile as three priority
groups with a score per CV. (b) One CV: the tool explains its score and asks
for a second look when it disagrees with the label. Both screens show the
synthetic pool.}
\label{fig:interface}
\end{figure}

Hiring for an academic position starts with a pile. One assistant professor
opening can draw well over a hundred applications, and a small committee has
to read all of them under a deadline. The order they are read in is usually
arbitrary: alphabetical, or by submission time, so strong candidates at the
bottom of the pile get read last by tired eyes.

Automating the decision is the wrong answer. Screening decisions are
consequential and legally sensitive, error costs are very unequal, and fully
automated rejection is the kind of system that has failed in public before
\citep{dastin2018amazon} and that current regulation treats as high risk
\citep{euaiact2024}. There is a weaker task that is both useful and much
safer: \emph{prioritization}. If a system proposes a good reading order, the
committee still reads everything and still makes every call, but the promising
files get attention first. The same framing has been standard for years in
technology-assisted review and in systematic-review screening
\citep{cormack2014evaluation, vandeschoot2021open}.

We have built and deployed such a tool. It runs at a national platform for
academic vacancies, integrated with that platform's vacancy database, and
Figure~\ref{fig:interface} shows what a committee actually gets from it. This
paper is not a system description; it reports what we found when we tried to
measure whether it works, and what we changed as a result. Four findings came
out of that. Each changed what we ship or what we claim, and none of them is
specific to our system.

\begin{enumerate}
  \item \textbf{Our evaluation collection was separable for a reason we
    did not intend.} Sorting the 300 synthetic CVs by character count puts every
    top-tier CV at the top of the list, because the generator wrote longer
    documents for stronger candidates. Precision at $k$ and nDCG on that pool
    cannot separate a relevance model from a word counter
    (Section~\ref{sec:pool}). Regenerating the pool under a shared length
    budget removes the shortcut and leaves the deployment claim standing
    (Section~\ref{sec:pool-v2}).
  \item \textbf{A four-level agreement statistic should not be read as a
    learning curve.}
    The one we report, like the one our interface shows a reviewer,
    discretizes a binary probability at fixed thresholds, so it moves with the
    classifier's confidence rather than with its ranking. Against held-out
    reference labels it rises to a peak and then falls as labels accumulate
    (Section~\ref{sec:metric}).
  \item \textbf{Labels are read twice.} In screening, the person who labels is
    the person who must read. Uncertainty sampling spends the budget on CVs a
    committee would otherwise skip, so it raises total screening effort even
    though the model it produces is fine (Section~\ref{sec:labels}).
  \item \textbf{The encoder only matters before the labels arrive.} A hosted
    LLM, used either as a scorer or as a drop-in replacement for the encoder,
    is far better than ours with no labels and indistinguishable from it after
    two dozen. A local cross-encoder gets most of the cold-start gain without
    sending a CV anywhere (Section~\ref{sec:llm}).
\end{enumerate}

\section{Related work}
\label{sec:related}

Ranking documents so that people find the relevant ones early is well studied
in systematic-review screening \citep{cohen2006reducing, wallace2010semi,
vandeschoot2021open} and in e-discovery \citep{cormack2014evaluation}, where
offline simulation over a labeled pool is the standard way to compare labeling
strategies \citep{ferdinands2023performance} and active learning is the
natural way to choose what to label next \citep{settles2009active}.
Decision-theoretic variants make the objective explicit, selecting by the
expected gain in a user-specified performance measure and under unequal
misclassification costs rather than by uncertainty alone
\citep{krempl2014probabilistic, krempl2015opal}; our result is about a cost
that such objectives do not usually include, namely the reading the labeling
itself creates. Uncertainty sampling in particular has a
long record of not beating random sampling in practice
\citep{attenberg2010inactive, lowell2019practical, munjal2022towards,
hacohen2022active}. CV screening has the same structure: a modest pool, a
binary decision downstream, asymmetric error costs, and a human who reads
everything that matters. Our contribution there is an accounting change rather
than another learning curve, since the annotator and the reader are the same
person.

Audits of commercial hiring systems find vague validity claims and little
transparency \citep{raghavan2020mitigating, bogen2018help, fabris2025fairness},
and the concern transfers to newer technology \citep{an2024hiring,
gaebler2025auditing, wilson2024gender}; our design responds by never filtering,
never training on historical outcomes, and showing the score and the
disagreements to the person who has to act on them, in line with a wider push
for transparency by design in applied NLP pipelines
\citep{mohammadi2024transparent, mohammadi2025explainability,
mohammadi2026explainnlp}. Building evaluation data with
language models is now common \citep{long2024synthetic, rahmani2024synthetic},
with unfaithfulness \citep{veselovsky2023faithful}, diversity loss
\citep{yu2023attributed} and uneven model-human agreement
\citep{mohammadi2025assessing} the documented risks; we add one that is easy to
miss and cheap to test for. Finally, how to summarize agreement is itself
contested, from chance correction \citep{cohen1968weighted,
mchugh2012interrater} to the behavior of summary scores under imbalance
\citep{chicco2020advantages, saito2015precision}; we report a wrinkle specific
to tools that show a person ordinal tiers cut from a binary probability.

\section{The deployed system}
\label{sec:system}

A reviewer pastes the job posting, uploads the CVs, labels a few, and reads
the resulting priority list; Figure~\ref{fig:system} in Appendix~\ref{app:params}
shows the pipeline underneath.\footnote{Public description of the tool for the
platform's member institutions:
\url{https://corporate.academictransfer.com/en/members/cv-priority-sorter/}.} The deployment context set the design, not
benchmark performance. The tool works per vacancy and per committee, because
relevance is local: a publication record that is outstanding for one chair is
beside the point for another. It has to learn from very few labels, since
nobody will annotate three hundred CVs to save time on three hundred CVs. It
must handle the platform's two working languages, Dutch and English. And it
has to be conservative with data.

The architecture is deliberately two-stage. A frozen multilingual transformer
supplies a 384-dimensional representation of every document, and a per-session
linear model learns one committee's preference on top of it. Nothing in the
encoder is fine-tuned; all adaptation happens in the head. That is what makes
retraining after every single label fast enough to feel interactive, and cheap
enough to discard when the session ends. A fine-tuned model could do neither,
and a model trained once on past hiring outcomes would import the bias this
design exists to avoid.

\paragraph{Processing and scoring.} Uploaded files are converted to text in
memory. Before any scoring, a redaction pass removes names, contact details,
addresses, birth dates and national identifiers, using pattern matching plus
named entity recognition. Institution names deliberately survive, since they
carry relevance signal. Each document is embedded with a multilingual MiniLM
sentence encoder \citep{reimers2019sentence, reimers2020making}, and the
posting is embedded the same way. Before any labels exist, a CV's score is a rescaled cosine similarity blended
with a keyword-match term, which orders the pile the moment the upload
finishes. Once the reviewer has labeled three relevant and three irrelevant
CVs, a logistic regression \citep{pedregosa2011scikit} is trained on the
document embedding plus its cosine similarity; while labels are few or one
class is rare, its probability is averaged with the zero-shot estimate. The
result is temperature-scaled \citep{guo2017calibration} and mapped through a
fixed piecewise-linear calibration onto a score from 0 to 100. Retraining takes
well under a second, so the list re-ranks after every label.
Appendix~\ref{app:params} lists every constant.

\paragraph{The loop with the reviewer.} Three interface decisions matter for
what follows. A \emph{suggest-next} button opens an unlabeled CV for the
reviewer to judge. When a label and the score disagree strongly, the tool
flags a \emph{conflict} and asks for a second look, as in
Figure~\ref{fig:interface}b; the reviewer can keep or change the label, and
either way the disagreement stays visible. And a self-check panel reports how
well the model currently reproduces the reviewer's own judgment, including a
four-level agreement coefficient. Nothing is persisted: the session, model
included, expires after 60 idle minutes or 8 hours.

\section{Evaluating it}
\label{sec:setup}

Realistic CV collections with ground truth are private, and using historical
application files requires ethical approval that only makes sense once a tool
has shown its behavior under controlled conditions. We therefore generated
300 synthetic academic CVs with a large language model against a real
assistant professor vacancy in computer science, in four relevance tiers of 75
each. A rubric over the six requirements of the posting defines the tiers, so
every CV has a reference label by construction, and the system under
evaluation never sees the rubric. Appendix~\ref{app:pool} gives the
construction.

We evaluate in two ways. \emph{Deployed-system runs} exercise the production
tool end to end: an operator uploads the pool, enters a prescribed subset of
the reference labels through the interface, and exports per-CV results. No
human judgment enters these runs, and they are not a user study.
\emph{Simulations} replicate the scoring pipeline offline with the same
encoder, features, classifier, blending and calibration, which lets us repeat
each condition over 20 seeds. A fixed stratified test set of 100 CVs is held
out and never available for labeling; every strategy and baseline is scored
on that same test set at every budget.

\paragraph{How well it works.} On the full pool the trained tool reaches
95.7\% binary accuracy against the reference labels, with errors confined to
adjacent tiers and both extreme tiers classified perfectly. The number that
matters more in deployment is reading depth: a committee reading in the tool's
order reaches 95\% of the relevant CVs after 49\% of the pile, against 63\%
with no labels at all and 95\% for an arbitrary order, where the best possible
on this pool is 47.5\%. Those are the numbers we would have reported before
the audit; Section~\ref{sec:pool} is why we now report only the second.

Two measurement choices matter. First, we report reading depth to 95\% recall
(and the work it saves, WSS@95, \citealp{cohen2006reducing}) rather than
precision at $k$, for the reason in Section~\ref{sec:pool}. Second, alongside
the linearly weighted $\kappa$ that the literature and our own interface use
\citep{cohen1968weighted}, we report Spearman's $\rho$ between the score and
the true tier, which needs no threshold; Section~\ref{sec:metric} explains
why that distinction turned out to matter more than we expected.

\section{What the audit found}

\subsection{The pool was sorted by length}
\label{sec:pool}

We had audited the pool for lexical diversity, checking that the generator had
not paraphrased itself into near-duplicates. That audit strips names,
institutions, dates and venue lists before measuring overlap, which removes
most of a CV's tier signal, so it cannot answer a second question: is the pool
separable for reasons that have nothing to do with relevance?

\begin{table}[t]
\centering
\small
\setlength{\tabcolsep}{2.5pt}
\begin{tabular}{lccccc}
\toprule
Ranker, no labels & nDCG & P@25 & P@75 & Read\,$\downarrow$ & $\rho$ \\
\midrule
Document length  & 0.994 & 1.00 & 1.000 & 0.640 & 0.862 \\
Line count       & 0.992 & 1.00 & 1.000 & 0.697 & 0.841 \\
Venue mentions   & 0.960 & 1.00 & 1.000 & 0.977 & 0.766 \\
BM25             & 0.991 & 1.00 & 0.987 & 0.723 & 0.805 \\
Zero-shot embed. & 0.918 & 0.72 & 0.787 & 0.653 & 0.634 \\
\bottomrule
\end{tabular}
\caption{Label-free rankers over the 300-CV pool. Read is the fraction of
the pile read to reach 95\% of the relevant CVs; $\rho$ is Spearman
correlation with the true tier.}
\label{tab:pool}
\end{table}

Table~\ref{tab:pool} answers the question directly. The generator wrote
longer CVs for stronger candidates: mean length runs from 3{,}945 characters
in the bottom tier to 6{,}292 in the top one, and the two are cut apart
entirely, since the shortest top-tier CV has 5{,}263 characters and the
longest CV in any other tier has 5{,}243. Sorting by character count therefore
fills the top 75 positions with the 75 top-tier CVs, at an nDCG of 0.994
against 0.9997 for our fully trained model. A second check makes the same
point from the supervised side: with five-fold cross-validation, TF-IDF over
the raw text reaches 0.950 four-tier accuracy against 0.893 for the deployed
encoder's embeddings, so this collection cannot show what a multilingual
sentence encoder adds. We keep it because the deployment has to handle two
languages that share little surface vocabulary.

What survives is the tail. Length ranks the top of the pool almost perfectly
and the bottom poorly, needing 64\% of the pile for 95\% recall against 49\%
for the trained model, because the last relevant CVs are the short ones. The
same holds for BM25. Reading depth to high recall is the measure the
deployment cares about, and it is the one this pool can still inform, so it is
the measure we lead with and the only one we treat as evidence.

The confound reaches less far than it might seem. It inflates
precision-style numbers and makes absolute quality here an upper bound, but it
does not touch the findings that follow: the calibration drift of
Section~\ref{sec:metric} is a property of a fixed threshold on a probability,
and the arithmetic of Section~\ref{sec:labels} turns on which CVs enter the
labeled set, not on how any of them rank. The lesson that transfers is that
a generator writes its label into surface features you did not ask it to use,
and a diversity audit will not find it. Rank a constructed collection with two or three model-free features before
trusting it.

\subsection{The repair, and what it reveals}
\label{sec:pool-v2}

The clean repair is to regenerate the pool so that length cannot carry the
label, and we have now done it. Pool v2 keeps the posting, the CV format and
the tier rubric verbatim and adds one instruction: every CV, whatever its
tier, is written to a budget of 700 to 800 words. Three things differ from
v1, for practical reasons: the generator is \texttt{gpt-5-mini}, since
the original model is no longer served; sampling runs at the default
temperature, with diversity supplied by per-candidate variation hints; and the
analysis is at text level, since v1's PDF rendering step is not reproducible
here. Out-of-budget drafts get one retry, and nothing is edited by hand. In
the end 90\% of the 300 CVs land in the accepted band, and the residual length-tier
correlation falls from 0.862 to 0.395, with tier ranges overlapping heavily
where v1's top tier was disjoint.

\begin{table}[t]
\centering
\small
\setlength{\tabcolsep}{2.5pt}
\begin{tabular}{lccccc}
\toprule
Ranker, no labels & nDCG & P@75 & Read\,$\downarrow$ & $\rho$ & $\rho$ len. \\
\midrule
Document length     & 0.955 & 0.827 & 0.847 & 0.395 & 1.000 \\
Venue mentions      & 0.996 & 1.000 & 0.477 & 0.934 & 0.544 \\
BM25                & 0.972 & 0.933 & 0.790 & 0.670 & 0.494 \\
Zero-shot embedding & 0.881 & 0.707 & 0.717 & 0.510 & 0.000 \\
Local cross-encoder & 0.982 & 1.000 & 0.477 & 0.912 & 0.300 \\
\bottomrule
\end{tabular}
\caption{Label-free rankers on pool v2, where every tier shares one length
budget. The optimum Read@95 for this pool is 0.475.}
\label{tab:v2}
\end{table}

The main result survives the repair. A session model trained on 24 random
labels reaches $\rho = 0.887$ and a reading depth of 0.482 on a held-out
quarter of v2, against an optimum of 0.475, so near-optimal reading depth
after a couple of dozen labels holds on a pool where length cannot help. The
cold start is where v2 differs (Table~\ref{tab:v2}). Without length to lean
on, the zero-shot encoder's discrimination falls from 0.634 to 0.510 and its
reading depth from 0.653 to 0.717, so part of its v1 score came from length.
The local cross-encoder of Section~\ref{sec:llm} closes that gap, reaching
0.477 with no labels at all. One more point matters to anyone building such a
collection: the pool did not become proxy-free, it changed proxies. Counting
mentions of the venues the posting names now separates the tiers almost
perfectly ($\rho = 0.934$), because the rubric is itself quantified in venues
and publication counts, and squeezing every tier into one length budget
concentrates that signal rather than diluting it. Controlling one proxy
promotes the next.

\subsection{An agreement score that is not a learning curve}
\label{sec:metric}

The four-level prediction our interface shows is not the output of a
four-class model. It is a binary probability pushed through a fixed
calibration curve and cut at fixed score thresholds, and those thresholds
never move. As the classifier grows more confident, its probability mass
slides across them, so the distribution of predicted tiers changes even when
the ordering does not.

\begin{figure}[t]
\centering
\includegraphics[width=\columnwidth]{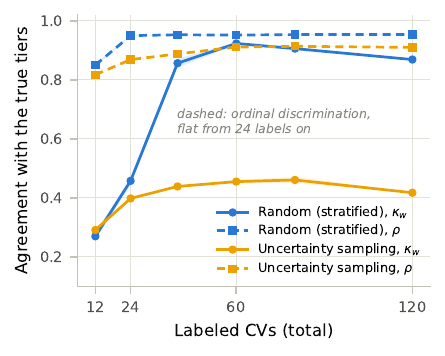}
\caption{The same runs under two statistics. Weighted $\kappa$ (solid) rises
and falls with the labeling budget; Spearman $\rho$ (dashed) is flat from 24
labels on.}
\label{fig:artifact}
\end{figure}

Figure~\ref{fig:artifact} shows the consequence. With labels drawn at random,
weighted $\kappa$ on the held-out test set rises from 0.270 at 12 labels to
0.924 at 60, then falls to 0.906 at 80 and 0.869 at 120. Read literally, the
model gets worse as the reviewer works harder. It does not. Ordinal
discrimination is flat from 24 labels on ($\rho$ of 0.950, 0.952 and 0.954 at
24, 60 and 120), and binary accuracy agrees, 0.913 at 24 labels against 0.925
at 120. What changes is confidence: mean $|p-0.5|$ goes from 0.117 at 24
labels to 0.251 at 120, and the fixed cut-points turn that into a moving tier
distribution, from 0/56/40/4 predicted at 24 labels to 31/24/17/29 at 120,
against a true 25/25/25/25.

The same mechanism accounts for a result we had read as substantive. Labeling
every available extreme CV and labeling every available middle CV differ by
0.039 in $\kappa_w$ (0.891 against 0.852) but by 0.005 in $\rho$ (0.954
against 0.949): almost the whole apparent difference between the two halves of
the difficulty range is calibration, not learning.

The point holds beyond our tool, because ordinal priority tiers
produced by thresholding a binary probability are common in deployed screening
tools: the underlying decision is binary, and the tiers exist to be shown to a
person. Whenever that is the design, an agreement statistic over those tiers
measures the model and a calibration curve nobody re-fits, and it should not be
plotted against labeling budget without a threshold-free companion.

\subsection{Labels are read twice}
\label{sec:labels}

The obvious way to build a suggest-next button is uncertainty sampling
\citep{lewis1994sequential}: open the unlabeled CV whose score is closest to
the decision boundary. Our first version did this. Given active
learning's record in deployment (Section~\ref{sec:related}), we simulated four
strategies over 20 seeds: stratified random, uniform random,
uncertainty sampling, and an ``extremes first'' habit that alternately labels
the highest- and lowest-scored unlabeled CV.

\begin{table}[t]
\centering
\small
\setlength{\tabcolsep}{3pt}
\begin{tabular}{lccc}
\toprule
Strategy (120 labels) & $\rho$\,$\uparrow$ & Relevant & Screened\,$\downarrow$ \\
\midrule
Random (stratified)  & 0.954 & 60.0 & 0.691 \\
Random (uniform)     & 0.954 & 61.0 & 0.689 \\
Uncertainty sampling & 0.910 & 22.1 & 0.817 \\
Extremes first       & 0.955 & 60.0 & 0.700 \\
\bottomrule
\end{tabular}
\caption{Labeling strategies at 120 labels, mean over 20 seeds. Relevant is
how many of the 120 labeled CVs were relevant; Screened is the share of the
pool read to reach 95\% recall, counting labeled CVs as read. The no-label
baseline is 0.653.}
\label{tab:strategies}
\end{table}

\begin{figure}[t]
\centering
\includegraphics[width=\columnwidth]{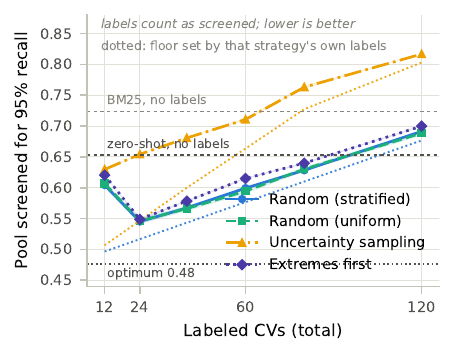}
\caption{Share of the pool screened to reach 95\% recall, by labeling
budget. Dotted lines are the floors set by the labeled sets of random
(stratified) and uncertainty sampling.}
\label{fig:simworkload}
\end{figure}

Model quality barely separates them. Three of the four sit within 0.005 of
each other in $\rho$ from 24 labels on, and uncertainty sampling, the worst,
trails by 0.04 (Table~\ref{tab:strategies}). Total effort separates them by an
order of magnitude more, and in the direction that costs the committee real
hours: uncertainty sampling ends at 0.817 of the pool against 0.691 for
random, and it is above the no-label baseline of 0.653 from two dozen labels
on, where the others stay below it until past 80
(Figure~\ref{fig:simworkload}).

The cause is arithmetic, not ranking. Over six budgets from 12 to 120 labels,
uncertainty sampling labels exactly one top-tier CV, and 82\% of what it
labels is irrelevant. Labeled CVs are screened by definition, and 95\%
recall means finding 143 of the 150 relevant CVs. A strategy that spends 120
readings and turns up only 22 of them still has to find 121 more, so no
ranker, however good, can screen less than 241 of the 300 CVs, or 0.803 of
the pool. Its measured 0.817 sits 0.014 above that floor, and random
labeling sits 0.014 above its own floor of 0.677. Both rankers are
essentially optimal on what is left. The entire difference is which CVs went
into the labeled set.

Two caveats belong with this result. Uncertainty sampling buys one thing:
it is the only strategy that never drops a relevant CV, where random labeling
leaves 3 to 4\% of the relevant test CVs behind, and that caution survives an
ablation of our blending rule. And part of the measured quality gap is our own
pipeline, not the strategy. The tool blends the model with the zero-shot score
whenever the labeled minority class is under 20\%, and uncertainty sampling's
label set is so one-sided that it never leaves that regime, while random
labeling leaves it at 24 labels. Scoring the identical label sets without the
blend closes most of the $\kappa_w$ gap, from 0.47 to 0.05 at 60 labels
(Table~\ref{tab:blend}). That
an internal safety rule can be held open indefinitely by a sampling strategy
is itself a deployment lesson. The workload result is untouched by it, because
it depends on the labeled set rather than on how the model is scored.

\subsection{What a bigger model buys, and when}
\label{sec:llm}

What would a current language model do instead of a 384-dimensional encoder
from 2020? We measured three candidates, none of which sees the tier rubric,
so the evaluation stays non-circular. As a \emph{scorer},
\texttt{gpt-5-mini} sees the posting and one CV and returns a match score from
0 to 100. As a \emph{vectorizer}, \texttt{text-embedding-3-small} (1536-d)
replaces our encoder and nothing else: same cosine zero-shot score, same
per-session regression, same calibration. As a \emph{local reranker}, a
multilingual cross-encoder (\texttt{jina-reranker-v2-base-multilingual}) reads
the posting and the CV together on the deployment's own hardware.

All three beat the deployed encoder at cold start, and they do it without
leaning harder on length. In a length-controlled band of 149 CVs the
vectorizer reaches $\rho=0.803$ and the cross-encoder 0.825 against our 0.659,
at length correlations near our own, and the hosted scorer reaches 0.874.
Table~\ref{tab:llm} gives the full pool as well.

The second half of the result decided what we do about it. Once the
per-session model has labels, the representation stops mattering. On the
held-out test set over 20 seeds the vectorizer gives $\rho$ of 0.949, 0.956
and 0.960 at 24, 60 and 120 labels against 0.950, 0.952 and 0.954 for the
deployed encoder, and reading depth of 0.526, 0.519 and 0.514 against 0.542,
0.535 and 0.530. A better representation is worth about 0.14 in $\rho$ before
the first label and 0.006 after two dozen: the head extracts what there is to
extract, and the encoder's real job is the part of the session where the
reviewer has told the system nothing.

The three price the upgrade differently. The scorer costs 262 seconds and
roughly 650k tokens per 300-CV pool, and both hosted variants send every CV to
a third party, which this deployment was built not to do. The cross-encoder
sends nothing out: it scored the pool in 76 seconds on a laptop, about 253 ms
per CV. Whichever is chosen, the upgrade buys the first two dozen CVs of a
session and almost nothing afterwards.

\subsection{What the redaction pass removes}
\label{sec:identity}

The tool redacts identifiers before scoring, so the last question was how much
of that machinery works. We rescored all 300 CVs with one fixed model.
Swapping the header name for names from four sets that differ in perceived
gender and ethnic association moves a CV's score by about 0.6 points out of
100 on average and moves at most one CV in or out of the top priority group,
close to the effect of changing the phone number. Masking affiliations, which deliberately survive
redaction, has about twice that effect and leaves reading depth unchanged
(Table~\ref{tab:identity}). Since the institutions here are invented, that is
a lower bound for a pool where prestige is real.

Because we know where the identifiers sit, the prior question is answerable
too: how much does the redaction pass remove? Over the 300 CVs it removes every
e-mail address, 87.9\% of phone numbers and, after the fix described below,
96.6\% of surnames. For names we count a leak whenever a token survives
anywhere in the text, in any casing, so the name figure is a lower bound on
recall. The exercise also found a defect: a middle
initial defeated name detection outright, because the stripped initial failed
the test that continues a name sequence, so ``Ada K.\ Lovelace'' yielded no
detected name at all.

\section{What we changed}
\label{sec:changed}

The audit changed the product as well as the paper. Two of the fixes below are
live in the deployed tool and the rest are recommendations: the name-detection
defect of Section~\ref{sec:identity}, for instance, was fixed while this paper
was written, moving surname removal from 92.3\% to 96.6\%.

The suggest-next button no longer implements uncertainty sampling. It now
rotates its suggestions across the four score bands, visiting the two middle
bands twice as often as the extremes, and it is deterministic in the session
state, so the suggestion does not change between renders; the guidance we give
reviewers is to label a couple of dozen CVs from across the range and stop. We
no longer report precision-style numbers on the synthetic pool. The priority
groups in the interface already fit their cut-points to the score distribution
of the pool at hand; the four-level tier bands behind the agreement score are
the ones that stay fixed, and moving them onto the reviewer's own labels is
the fix Section~\ref{sec:metric} asks for. The remaining recommendation is
that a fallback rule like our low-confidence blend should report when it is
active, since a sampling strategy can hold one open indefinitely without
anyone noticing. The encoder is the one open decision. Section~\ref{sec:llm} prices the
upgrade: the hosted variants would need a privacy arrangement, while the local
cross-encoder needs none, so for the cold start what remains is engineering
work.

\section{Conclusion}

A prioritization tool is a modest thing to build and harder to measure well
than it looks. The findings we report are not about modeling: an evaluation
collection that encoded its labels in document length, an agreement statistic
that followed calibration rather than ranking, a labeling strategy that
improved nothing and left the committee reading more of the pile than with no
labels at all, and an encoder whose choice turns out to matter only until the
first labels arrive. Each came from re-checking something we had already
trusted, and each transfers to any team that builds its own screening collection, reports a
tiered agreement score, or asks a reviewer what to label next. The first three
changed the product; the fourth prices the next change.

\section*{Limitations}

Everything here is measured on synthetic CVs, so the difficulty of the task is
set by the generator rather than by the world. Beyond the length confound of
Section~\ref{sec:pool}, this means absolute quality on this pool is an upper
bound on what the same pipeline would achieve in the field, and comparisons
between feature sets on it are unreliable. The regeneration under one shared
length budget (Section~\ref{sec:pool-v2}) removes the length shortcut and the
deployment claim survives it, but the new pool can be sorted by venue mentions
instead, and the same caution applies to it.

Balanced prevalence is a second distortion. Half of this pool is relevant, and
reading depth to 95\% recall has a ceiling that depends directly on that
fraction, so ``half the pile'' does not transfer; the same sensitivity is a
known problem for WSS in citation screening \citep{kusa2023analysis}.
Re-evaluating a fixed trained model on subsamples at lower prevalence shows
both sides of it: at 10\% relevance the committee reaches 95\% recall after
16\% of the pile, which is better in absolute terms, but the distance to the
achievable optimum widens from 1.06$\times$ to 1.63$\times$.

The pool is built from a single vacancy in a single field, where requirements
are unusually legible: named venues and countable publications. Reference
labels are one-dimensional, while real committees weigh criteria that
interact. The simulations replicate the production pipeline faithfully but not
perfectly, since the offline zero-shot score omits the keyword-matching
component; that also means our blend ablation overstates what the blending
rule costs. The 20 seeds are not equally independent across strategies:
two random runs share 29\% of their labels at a budget of 60, two
uncertainty-sampling runs share 96\%, so paired tests against the sequential
strategies carry less weight than the seed count suggests.

The LLM comparison is one model of each kind, on one pool, with one prompt;
prompt sensitivity is not something we measured, so it bounds what the upgrade
is worth here rather than settling how these families compare in general,
and the cross-encoder's speed was measured on a laptop rather than on the
deployment's servers. The
perturbation probe and the redaction figures describe the deployed MiniLM
pipeline, and neither transfers to a representation they never saw: if the
encoder is replaced, both have to be re-run. Finally, no human judgment
enters this evaluation at all. We measure prioritization quality, not reading
time, not agreement between committee members, and not whether a proposed
order is one anybody would follow.

\section*{Ethics Statement}

No real applicant data was used in this work. All 300 CVs are synthetic, with
fictional names and contact details, generated for this study; the one real
document involved is a public job posting. The name-perturbation experiment in
Section~\ref{sec:identity} substitutes names into synthetic documents to
measure a pipeline's sensitivity, and no claim is made that a name determines
any attribute of a real person. The LLM experiments in Section~\ref{sec:llm}
sent only these synthetic documents to a hosted API; the deployed tool itself
calls no third-party service.

The deployed tool is designed around data minimization: in-memory processing,
redaction of personal identifiers before scoring, session expiry, and no
third-party services. It recommends a reading order and takes no decision
about any candidate, so the worst failure mode is a suboptimal reading order
rather than a rejection. It never trains on historical hiring outcomes, which
is the failure mode of \citet{dastin2018amazon}. Redaction is not a fairness
guarantee, and proxies survive it; because the system deliberately never
learns identity attributes, outcome monitoring such as the adverse impact
ratio \citep{roth2006modeling} or exposure-based fair-ranking measures
\citep{zehlike2022fairness} has to happen outside the scoring pipeline, on the
committee's side, with the exported ranking. Under the EU AI Act, systems used
in employment decisions fall in the high-risk class \citep{euaiact2024}; a
prioritization tool with a human reading every file sits at the mild end of
that class, and the obligations it anticipates are reflected in the interface
shown in Figure~\ref{fig:interface}. A pilot with anonymized historical
applications is planned separately, conditional on ethical approval.

\section*{Acknowledgments}
This work was supported by a grant from the Applied Data Science focus area
of Utrecht University and by AcademicTransfer.

\bibliography{references}

\appendix

\section{Scoring pipeline parameters}
\label{app:params}

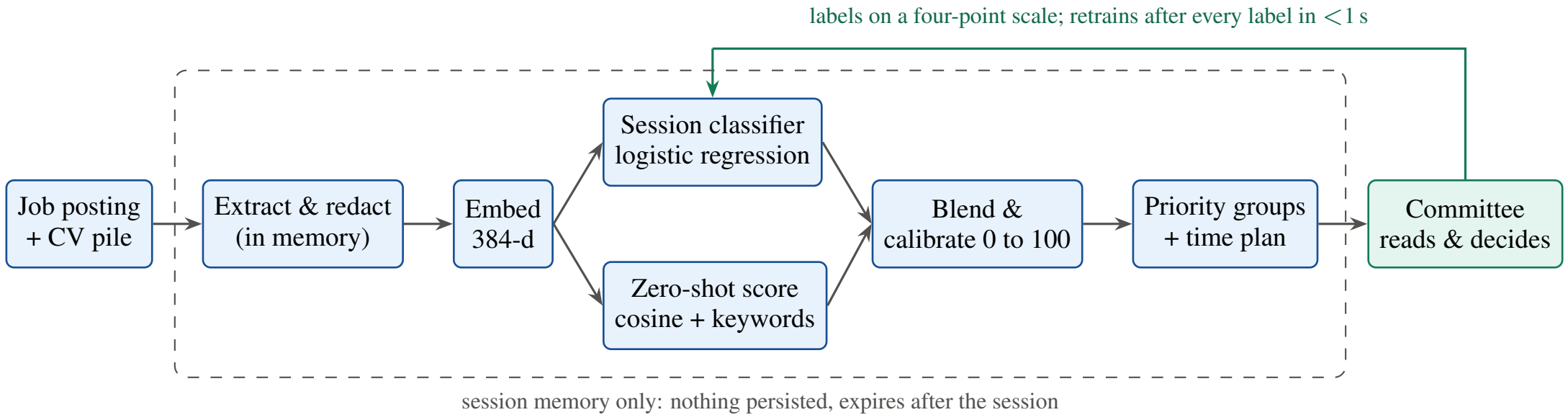
\begin{figure*}[t]
\centering
\begin{tikzpicture}[
  font=\scriptsize,
  box/.style={draw=seqdark, fill=seqlight!45, rounded corners=2pt,
              line width=0.5pt, align=center, inner sep=3pt,
              minimum height=8mm},
  human/.style={draw=accentaqua!70!black, fill=accentaqua!12,
              rounded corners=2pt, line width=0.5pt, align=center,
              inner sep=3pt, minimum height=8mm},
  arr/.style={-{Stealth[length=1.6mm]}, line width=0.6pt, draw=inkgray},
]
\node[box] (input) {Job posting\\+ CV pile};
\node[box, right=4.5mm of input] (redact) {Extract \& redact\\(in memory)};
\node[box, right=4.5mm of redact] (embed) {Embed\\384-d};
\node[box, right=4.5mm of embed, yshift=7.5mm]
  (clf) {Session classifier\\logistic regression};
\node[box, right=4.5mm of embed, yshift=-7.5mm]
  (zeroshot) {Zero-shot score\\cosine + keywords};
\node[box, right=4.5mm of clf, yshift=-7.5mm] (blend)
  {Blend \&\\calibrate 0 to 100};
\node[box, right=4.5mm of blend] (groups) {Priority groups\\+ time plan};
\node[human, right=4.5mm of groups] (committee) {Committee\\reads \& decides};
\draw[arr] (input) -- (redact);
\draw[arr] (redact) -- (embed);
\draw[arr] (embed.east) -- (zeroshot.west);
\draw[arr] (embed.east) -- (clf.west);
\draw[arr] (zeroshot.east) -- (blend.west);
\draw[arr] (clf.east) -- (blend.west);
\draw[arr] (blend) -- (groups);
\draw[arr] (groups) -- (committee);
\draw[line width=0.6pt, draw=accentaqua!70!black]
  (committee.north) |-
  node[pos=0.75, above=0.5mm, font=\tiny, text=accentaqua!60!black]
  {labels on a four-point scale; retrains after every label in $<$1\,s}
  ([yshift=4.5mm]clf.north);
\draw[arr, draw=accentaqua!70!black]
  ([yshift=4.5mm]clf.north) -- (clf.north);
\node[draw=inkgray, dashed, rounded corners=3pt, line width=0.4pt,
      inner sep=2.5mm, fit=(redact) (embed) (zeroshot) (clf) (blend) (groups),
      label={[font=\tiny, text=inkgray]below:%
      session memory only: nothing persisted, expires after the session}]
  {};
\end{tikzpicture}
\caption{The pipeline. A zero-shot score orders the pile at upload; a
per-session classifier takes over as labels arrive. Nothing inside the dashed
boundary outlives the session.}
\label{fig:system}
\end{figure*}

Encoder \texttt{paraphrase-multi\-lingual-\allowbreak MiniLM-L12-v2}, 384
dimensions.
Documents are split into 500-character chunks, at most 16 per document; chunk
vectors are mean-pooled and L2-normalized. The posting is embedded as
$0.6\,\mathbf{e}_{\text{full}} + 0.4\,\mathbf{e}_{\text{req}}$, renormalized.
The zero-shot semantic component rescales the cosine similarity as
$\mathrm{clip}((\cos-0.25)/0.50, 0, 1)$; in production this is blended with a
keyword-match term that the offline replica omits. The session classifier is a
scikit-learn logistic regression, $C=1.0$, balanced class weights, at most
1000 iterations, on 385 features. Training starts once at least three labels
exist in each binary class. While the labeled set holds fewer than 20 items,
or its minority class holds under 20\% of them, the reported probability is
$0.5\,p_{\text{model}} + 0.5\,p_{\text{zero-shot}}$. Calibration is
temperature scaling with $T=0.85$ followed by a piecewise-linear map with
breakpoints $0\!\to\!0$, $0.20\!\to\!20$, $0.45\!\to\!45$, $0.70\!\to\!70$,
$0.85\!\to\!82$, $1\!\to\!99$; tier cut-points on the resulting score are
$\geq 75$, $\geq 50$ and $\geq 26$. These cut-points are fixed at build time,
which is the mechanism analyzed in Section~\ref{sec:metric}.

\section{The synthetic pool}
\label{app:pool}

We generated 300 CVs with \texttt{gpt-5.4-mini} at temperature 2, one per
request. The prompt fixes three parts: the full text of a real assistant
professor vacancy in computer science, whose six requirements cover the PhD,
publication venues, systems building, teaching, funding and communication; an
eleven-part academic CV structure; and a rubric defining four tiers against
those requirements. Top-tier CVs satisfy all six requirements with more than
six publications, at least three in the venues the posting names. The second
tier satisfies about four, with exactly three publications, one top venue, and
a PhD outside computer science. The third satisfies about one, with a single
non-top-venue publication. The fourth satisfies none. Each tier holds 75 CVs,
rendered to PDF so the pool exercises the real document pipeline. Lexical
overlap within tiers, measured on 5-gram shingles with names, institutions,
dates and venue lists stripped, runs from 9.7\% to 20.6\%; cross-tier overlap
stays at or below 6.2\%. Section~\ref{sec:pool} explains why that audit is not
sufficient.

\section{Additional results}
\label{app:extra}

\begin{table}[h]
\centering
\small
\setlength{\tabcolsep}{3pt}
\begin{tabular}{lccc}
\toprule
No labels & $\rho$\,$\uparrow$ & $\rho$ vs len. & Read@95\,$\downarrow$ \\
\midrule
\multicolumn{4}{l}{\emph{full pool}} \\
LLM scorer            & 0.955 & 0.863 & 0.580 \\
LLM vectorizer        & 0.894 & 0.813 & 0.567 \\
Local cross-encoder   & 0.937 & 0.869 & 0.560 \\
Deployed encoder      & 0.634 & 0.596 & 0.653 \\
\addlinespace
\multicolumn{4}{l}{\emph{length-controlled band, $n=149$}} \\
LLM scorer            & 0.874 & 0.488 & 0.617 \\
LLM vectorizer        & 0.803 & 0.499 & 0.564 \\
Local cross-encoder   & 0.825 & 0.540 & 0.577 \\
Deployed encoder      & 0.659 & 0.481 & 0.671 \\
\bottomrule
\end{tabular}
\caption{Cold start, before any labels. The middle column is the correlation
with document length, the confound of Section~\ref{sec:pool}.}
\label{tab:llm}
\end{table}

\begin{table}[h]
\centering
\small
\setlength{\tabcolsep}{4pt}
\begin{tabular}{lcccc}
\toprule
Perturbation & $\tau$ & $\overline{|\Delta s|}$ & Top-75 & Read \\
\midrule
Phone number (control)   & 0.974 & 0.49 & 0 & 0.500 \\
Name $\rightarrow$ placeholder & 0.970 & 0.62 & 1 & 0.503 \\
Name set A & 0.969 & 0.59 & 1 & 0.503 \\
Name set B & 0.970 & 0.57 & 1 & 0.500 \\
Name set C & 0.971 & 0.59 & 0 & 0.500 \\
Name set D & 0.967 & 0.66 & 1 & 0.500 \\
Affiliations masked & 0.956 & 0.96 & 2 & 0.500 \\
\bottomrule
\end{tabular}
\caption{Ranking movement under identity perturbations. $\tau$ is Kendall
correlation with the unperturbed ranking; $\overline{|\Delta s|}$ is the mean
absolute score change on the 0 to 100 scale; Top-75 counts CVs entering or
leaving the top group; Read is reading depth for 95\% recall.}
\label{tab:identity}
\end{table}

\begin{table}[h]
\centering
\small
\setlength{\tabcolsep}{2.5pt}
\begin{tabular}{lcccccc}
\toprule
Budget & 12 & 24 & 40 & 60 & 80 & 120 \\
\midrule
\multicolumn{7}{l}{\emph{$\kappa_w$ as reported}} \\
Random (strat.)  & 0.270 & 0.458 & 0.857 & 0.924 & 0.906 & 0.869 \\
Uncertainty      & 0.291 & 0.398 & 0.439 & 0.455 & 0.461 & 0.418 \\
\addlinespace
\multicolumn{7}{l}{\emph{$\kappa_w$, blend disabled at scoring time}} \\
Random (strat.)  & 0.401 & 0.458 & 0.857 & 0.924 & 0.906 & 0.869 \\
Uncertainty      & 0.286 & 0.473 & 0.829 & 0.878 & 0.862 & 0.805 \\
\addlinespace
\multicolumn{7}{l}{\emph{Mean pairwise overlap of the 20 labeled sets}} \\
Random (strat.)  & 0.060 & 0.118 & 0.189 & 0.294 & 0.398 & 0.601 \\
Uncertainty      & 0.763 & 0.855 & 0.915 & 0.958 & 0.978 & 0.988 \\
\bottomrule
\end{tabular}
\caption{Robustness of the comparison in Section~\ref{sec:labels}. Random
labeling is blended only at a budget of 12, so its row changes in one column;
uncertainty sampling is blended at every budget.}
\label{tab:blend}
\end{table}

\end{document}